# Principal Basis Analysis in Sparse Representation

Hong Sun*[a,b], Cheng-Wei Sang[a], Chen-Guang Liu[a]

[a]School of electronic Information, Wuhan University, Wuhan, 430072, China;
[b]Telecom ParisTech, 46, rue Barrault, 75013 Paris, France

**ABSTRACT**

This article introduces a new signal analysis method, which can be interpreted as a principal component analysis in sparse decomposition of the signal. The method, called principal basis analysis, is based on a novel criterion: reproducibility of component which is an intrinsic characteristic of regularity in natural signals. We show how to measure reproducibility. Then we present the principal basis analysis method, which chooses, in a sparse representation of the signal, the components optimizing the reproducibility degree to build the so-called principal basis. With this principal basis, we show that the underlying signal pattern could be effectively extracted from corrupted data. As illustration, we apply the principal basis analysis to image denoising corrupted by Gaussian and non-Gaussian noises, showing better performances than some reference methods at suppressing strong noise and at preserving signal details.

## 1. INTRODUCTION

Principal component analysis (PCA) [1] is a powerful tool in signal analysis knowing that it is a common technique for seeking underlying signal patterns in corrupted data set with numerous applications to denoising, or inverse problems, also widely used in feature extraction, or machine learning [2], [3]. The goal of PCA is to identify the most meaningful basis to project a data set, and this basis will hopefully filter out the noise and reveal hidden structure. In this section, some limitations of PCA in applications will be seen and our new principal components analysis in sparse representation of the signal, called as principal basis analysis, will be proposed to strengthen signal analysis capabilities.

Consider a data matrix, $\mathbf{X} = \left[ \{x_{i,j}\}_{1 \leq i \leq N, 1 \leq j \leq M} \right] \in \mathbb{R}^{N \times M}$, where each of the $N$ rows represents a different repetition of the experiment, and each of the $M$ columns gives a particular kind of datum $\mathbf{x}_m = \{x_{i,m}\}_{i=1}^N \in \mathbb{R}^N$, $1 \leq m \leq M$.

The PCA decomposition of $\mathbf{X}$ can be obtained from the singular value decomposition (SVD):

$$\mathbf{X}_{N \times M} = \mathbf{U}_{N \times N} \mathbf{A}_{N \times M} = \mathbf{U}_{N \times N} \mathbf{\Sigma}_{N \times M} \mathbf{V}^T_{M \times M} \qquad (1)$$

where $\mathbf{U}^T\mathbf{U} = \mathbf{I}_N, \mathbf{V}^T\mathbf{V} = \mathbf{I}_M$ ($\mathbf{I}$ identity matrix) and $\mathbf{\Sigma} = \begin{bmatrix} diag[\sigma_1, \cdots, \sigma_r], \mathbf{0} \\ \mathbf{0} \end{bmatrix}$, $\{\sigma_i\}_{i=1}^r$ are positive real and termed the singular values for the symmetric matrix $\mathbf{X}^T\mathbf{X}$ with a rank $r(r \leq \min[N, M])$. Eqn.1 can be re-written in terms of vectors as:

$$[\mathbf{x}_1, \cdots, \mathbf{x}_m, \cdots, \mathbf{x}_M] = [\mathbf{u}_1, \cdots, \mathbf{u}_n, \cdots, \mathbf{u}_N] \bullet [\boldsymbol{\alpha}_1, \cdots, \boldsymbol{\alpha}_m, \cdots, \boldsymbol{\alpha}_M]^T \qquad (2)$$

where $\mathbf{u}_n \in \mathbb{R}^N$ and $\boldsymbol{\alpha}_m \in \mathbb{R}^{1 \times N}$. Eqn.2 means that the data set $\{\mathbf{x}_m\}_{m=1}^M$ is expressed on the orthonormal basis $\{\mathbf{u}_n\}_{n=1}^N$ as $\{\boldsymbol{\alpha}_m\}_{m=1}^M$.

Hong SUN is with School of Electronic Information, Wuhan University, 430072 Wuhan, China, and Signal and Image Processing Department, Telecom ParisTech, 46 rue Barrault, 75013 Paris, France. hongsun@whu.edu.cn;
Cheng-Wei SANG and Chen-Guang LIU are with School of Electronic Information, Wuhan University, Wuhan 430072, China

In PCA decomposition (Eqn.1), standard variation $\sigma_i$ is used as the measurement for identifying the importance of vector $\mathbf{u}_i$. With this score parameter, a truncation procedure is generally explored in applications: Reordering $\{\sigma_i\}_{i=1}^r$ to become $\{\sigma_i'\}_{i=1}^r$ where $\sigma_1' \geq \sigma_2' \geq \cdots \geq \sigma_r'$, denoting the sorted diagonal matrix $\mathbf{\Sigma}' = \begin{bmatrix} diag[\sigma_1', \cdots, \sigma_r'], \mathbf{0} \\ \mathbf{0} \end{bmatrix}$ and its accompanying orthogonal matrices $\mathbf{U}'$ and $\mathbf{A}'$, Eqn.2 can be re-written as:

$$[\mathbf{x}_1, \cdots, \mathbf{x}_m, \cdots, \mathbf{x}_M] = \mathbf{U}'\mathbf{A}' = [\mathbf{u}_1', \cdots, \mathbf{u}_n', \cdots, \mathbf{u}_N'] \bullet [\boldsymbol{\alpha}_1', \cdots, \boldsymbol{\alpha}_m', \cdots, \boldsymbol{\alpha}_M']^T \quad (3)$$

Thus, the first $P(P<r)$ components $\{\mathbf{u}_n'\}_{n=1}^P$ are considered as the most meaningful feature vectors.

For many applications of PCA, the truncation of a matrix $\mathbf{U}_T' = \{\mathbf{u}_n'\}_{n=1}^P$ or $\mathbf{A}_T' = \{\boldsymbol{\alpha}_m' \in \mathbb{R}^{1 \times P}\}_{m=1}^M$ produces an optimal underlying signal pattern $\hat{\mathbf{S}}$ in the sense of the smallest possible Frobenius norm of difference between $\hat{\mathbf{S}}$ and the true signal $\mathbf{S}$ having. This PCA truncation supposes that the components of the true signal behind the data set have a maximum variance and the other components are mainly due to noise. However, in many practical cases, some components with low variance might actually be important relative to signal details, and some types of noise with non-Gaussian statistics noise, might actually have a significant variance. Besides, it is difficult to give the orthonormal basis $\{\mathbf{u}_i\}_{i=1}^r$ a physical interpretation although it has very good characteristics in the mathematical sense. Sometimes, the best way to represent a signal is not as a combination of additive components. It is for instance the case in image processing where state of objects is of prime importance.

In recent years, a growing interest in research has been put on sparse representation of a data set $\mathbf{X} = \{\mathbf{x}_m \in \mathbb{R}^N\}_{m=1}^M \in \mathbb{R}^{N \times M}$ with over-complete dictionary $\mathbf{D} = \{\mathbf{d}_k\}_{k=1}^K \in \mathbb{R}^{N \times K} (K > N)$ where $\mathbf{d}_k \in \mathbb{R}^N$ is a prototype signal pattern, called as atom [4]. The sparse decomposition of $\mathbf{X}$ can be given as:

$$\mathbf{X}_{N \times M} = \mathbf{D}_{N \times K} \mathbf{A}_{K \times M} = [\mathbf{d}_1, \cdots, \mathbf{d}_k, \cdots, \mathbf{d}_K] \bullet [\boldsymbol{\alpha}_1, \cdots, \boldsymbol{\alpha}_m, \cdots, \boldsymbol{\alpha}_M]^T \quad (4)$$

with sparse coefficient matrix $\mathbf{A} = \{\alpha_{k,m}\}_{\substack{1 \leq k \leq K \\ 1 \leq m \leq M}} = \{\boldsymbol{\alpha}_m\}_{m=1}^M$ where row vector $\boldsymbol{\alpha}_m \in \mathbb{R}^{1 \times N}$. By learning from examples, $K$ optimal dictionary $\{\hat{\mathbf{d}}_k\}_{k=1}^K \triangleq \hat{\mathbf{D}}$ and their sparse coefficient matrix $\{\hat{\boldsymbol{\alpha}}_m\}_{m=1}^M \triangleq \hat{\mathbf{A}}$ can be obtained by solving $M$ forms:

$$\{\hat{\mathbf{D}}, \hat{\boldsymbol{\alpha}}_m\} = \arg\min_{\mathbf{D}, \boldsymbol{\alpha}_m} \|\boldsymbol{\alpha}_m\|_0 + \|\mathbf{D}\boldsymbol{\alpha}_m - \mathbf{x}_m\|_2^2 \leq \varepsilon, \quad 1 \leq m \leq M \quad (5)$$

where $\varepsilon$ is allowed error tolerance. A typical way to solve Eqn.5 is for instance K-SVD algorithm [6]. The first term in Eqn.5 constrains the solution with the fewest number of nonzero coefficients in each of sparse coefficient vectors $\boldsymbol{\alpha}_m (1 \leq m \leq M)$. The underlying assumption of sparse decomposition (Eqns.4 and 5) expresses the widely known fact that natural signals could be represented by combining few components only. Typically, an optimal approximation of the underlying signal $\mathbf{S}$ from corrupted data set $\mathbf{X}$ is obtained as $\mathbf{S} = \hat{\mathbf{D}}\hat{\mathbf{A}}$ without any truncating process.

A question here arises naturally: why not produce a principal basis by choosing the most meaningful components from optimal dictionary $\hat{\mathbf{D}}$ to improve efficiency of the sparse decomposition? Maybe we have not yet a suitable score parameter (like $\sigma$ in PCA decomposition as Eqn.1) to measure significance of prototype component or atom $\hat{\mathbf{d}}_k$ since basis $\{\mathbf{d}_k\}_{k=1}^K$ is non-orthogonal and non-complete.

We propose a new concept to access the significance of atoms in sparse decomposition with an intrinsic feature of natural signal; and propose a measurement index of the atom significance in Section II. Then we present a method, called Principal Basis Analysis (PBA), to build a principal basis consisting of the most meaningful atoms and to produce an optimal approximation of the underlying signal with this principal basis in Section III. Some applications of the proposed PBA method to denoising are shown and some comparisons are made in Section IV.

## 2. IMPORTANCE RANK OF ATOMS

As previously stated, PCA decomposition (Eqn.1) constrains bases $\mathbf{U}$ or $\mathbf{V}$ to be orthonormal, consequently, the score matrix $\mathbf{\Sigma}$ consists of a diagonal matrix of standard variation $\sigma_i$ which indicates the strength of component vector $\mathbf{u}_i$. In comparison, sparse decomposition (Eqn.4) takes large number of signal pattern $\mathbf{d}_k$ as component with sparse constraint, which means a widely known fact that natural signals could be combined with few principal components. Or, to put it another way, the sparse assumption of coefficient vectors (Eqn.5) means that those few components (atoms) are frequently occurrent in signal sparse representation. In other words, sparse coefficient presentation means high reproducibility of the atoms. Intuitively, high reproducibility of signal components is actually an intrinsic characteristic of the signal regularity, such as sharp transitions, singularities, textures, smooth transitions and so on. On the contrary, low reproducibility of components is a more general characteristic for many different types of noise, Gaussian or non-Gaussian.

Now, we have just cause for proposing a measurement to assess atom importance under the concept of reproducibility. Consider the sparse coefficient matrix $\hat{\mathbf{A}}_{K \times M}$ in Eqn.4 in terms of column vectors $\boldsymbol{\lambda}_k = \{\hat{\alpha}_{k,m}\}_{m=1}^M \in \mathbb{R}^{1 \times M}$ instead of row vectors $\hat{\boldsymbol{\alpha}}_m$:

$$\hat{\mathbf{A}} = \{\hat{\alpha}_{k,m}\}_{1 \le k \le K, 1 \le m \le M} = \left[\{\boldsymbol{\lambda}_k\}_{k=1}^K\right]^T \tag{6}$$

Thus the sparse representation of data set $\mathbf{X}$ can be written as

$$\mathbf{X} \approx \hat{\mathbf{D}}\hat{\mathbf{A}} = \left[\hat{\mathbf{d}}_1, \cdots, \hat{\mathbf{d}}_k, \cdots, \hat{\mathbf{d}}_K\right] \bullet [\boldsymbol{\lambda}_1, \cdots, \boldsymbol{\lambda}_k, \cdots, \boldsymbol{\lambda}_K]^T \tag{7}$$

Eqn.7 tells us that the vector $\boldsymbol{\lambda}_k$ reflects characteristics of the component $\hat{\mathbf{d}}_k$. What is most interesting is its $\ell^0$ zero pseudo-norm $\|\boldsymbol{\lambda}_k\|_0$ which counts the non-zero entries, which means the occurrence frequency of $\hat{\mathbf{d}}_k$ in the sparse representation. It actually happens that $\|\boldsymbol{\lambda}_k\|_0$ is just right a simple measurement as an index of the reproducibility of the component $\hat{\mathbf{d}}_k$ for data $\mathbf{X}$.

To illustrate that reproducibility criterion is more robust for distinguishing signal details from noise than commonly used strength criterion, we show an image example and its $\ell^0$-norms $\|\boldsymbol{\lambda}_k\|_0, \forall k$ interpreted as reproducibility of components $\hat{\mathbf{d}}_k$ s and $\ell^1$-norms $\|\boldsymbol{\lambda}_k\|_1, \forall k$ interpreted as strength of components $\hat{\mathbf{d}}_k$ s. Taking a natural image of size $64 \times 64$ with some details and a Gaussian white noise image of the same size with standard deviation $\sigma = 35$, their over-complete dictionaries $\hat{\mathbf{D}}$ with 256 atoms (each patch of size $8 \times 8$) and the $\ell^0$-norms and $\ell^1$-norms of their sparse coefficient vectors $\{\boldsymbol{\lambda}_k \in \mathbb{R}^{1\times 64}\}_{k=1}^{256}$ (Eqn.6) obtained by solving Eqn.5 are shown in Fig.1. We can see that the $\ell^0$-norms of $\boldsymbol{\lambda}$ s of the natural image $\{\|\boldsymbol{\lambda}_k^{image}\|_0\}_{k=1}^{256}$ (red lines) are much larger, except the null points, than those of noise $\{\|\boldsymbol{\lambda}_k^{noise}\|_0\}_{k=1}^{256}$ (black lines), which means that components in noise decomposition have normally very low reproducibility. Looking at the $\ell^1$-norms of $\boldsymbol{\lambda}$ s, many $\|\boldsymbol{\lambda}_k^{image}\|_1$ (red lines) of the natural image corresponding to components of faint details are smaller than those of the noise $\|\boldsymbol{\lambda}_k^{noise}\|_1$ (black lines) and some of noise $\|\boldsymbol{\lambda}_k^{noise}\|_1$ corresponding to strong noise are rather large.

Analytically and experimentally, we propose taking the $\ell^0$-norm $\|\boldsymbol{\lambda}_k\|_0$ of column vector of sparse coefficient matrix $\hat{\mathbf{A}}$ as index to measure reproducibility of the components $\hat{\mathbf{d}}_k$ in sparse representation (Eqn.7) of data set $\mathbf{X} = \{\mathbf{x}_m\}_{m=1}^M$.

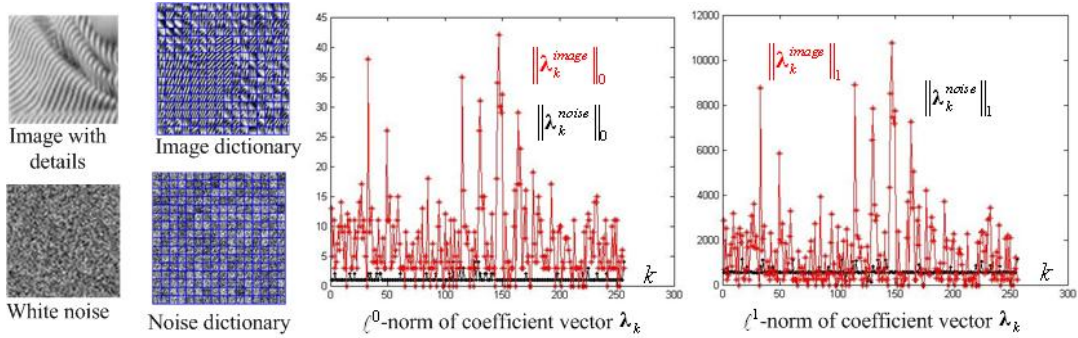

Figure 1. Component reproducibility measured by the zero pseudo-norm of coefficient vector and strength measured by the first norm for signal details (in red) and white noise (in black).

## 3. PRINCIPAL BASIS ANALYSIS

Instead of the principal components analysis (PCA) taking the eigenvectors with larger eigenvalue in the eigen-decomposition as principal signal components [4], we propose here a principal basis analysis (PBA) taking the atoms with larger $\ell^0$-norms of coefficient vectors in the sparse representation as principal signal basis.

Considering signal matrix $\mathbf{X} \in \mathbb{R}^{N\times M}$, we get an over-complete dictionary $\hat{\mathbf{D}} \in \mathbb{R}^{N\times K}$ with $K > N$ and its sparse coefficient matrix $\hat{\mathbf{A}} \in \mathbb{R}^{K\times M}$ subject to $\hat{\mathbf{D}}\hat{\mathbf{A}} \approx \mathbf{X}$ (Eqn.5). Taking the vectors $\{\boldsymbol{\lambda}_k\}_{k=1}^K$ from $\hat{\mathbf{A}}$ (Eqn.6). We calculate the $\ell^0$-norms $\{\|\boldsymbol{\lambda}_k\|_0\}_{k=1}^K$ and rank them in descending order:

$$\boldsymbol{\Lambda} \triangleq [\boldsymbol{\lambda}_1', \cdots, \boldsymbol{\lambda}_k', \cdots, \boldsymbol{\lambda}_K'] \overset{sort}{\Leftarrow} [\boldsymbol{\lambda}_1, \cdots, \boldsymbol{\lambda}_k, \cdots, \boldsymbol{\lambda}_K] \text{ where } \|\boldsymbol{\lambda}_1'\|_0 \geq \|\boldsymbol{\lambda}_2'\|_0 \geq \cdots \geq \|\boldsymbol{\lambda}_K'\|_0 \qquad (8)$$

With reordered atoms $\{\mathbf{d}_k'\}_{k=1}^K$ corresponding to $\{\|\boldsymbol{\lambda}_k'\|_0\}_{k=1}^K$, the reordered dictionary $\mathbf{D}'$ is written as:

$$\hat{\mathbf{D}} \overset{reorder}{\Rightarrow} \mathbf{D}' = [\mathbf{d}_1', \cdots, \mathbf{d}_k', \cdots, \mathbf{d}_K'] \qquad (9)$$

So we have a sparse representation of data as $\mathbf{X} \approx \hat{\mathbf{D}}\hat{\mathbf{A}} = \mathbf{D}'\mathbf{\Lambda}$.

Thereupon, we take the first $P$ atoms $\mathbf{d}'_1, \mathbf{d}'_2, \cdots, \mathbf{d}'_P$ to form a principal dictionary $\overline{\mathbf{D}}_P \in \mathbb{R}^{P \times N}$ as a principal signal basis:

$$\overline{\mathbf{D}}_P = truncating_P[\mathbf{D}'] = [\mathbf{d}'_1, \mathbf{d}'_2, \cdots, \mathbf{d}'_P] \tag{10}$$

The remaining atoms $\mathbf{d}'_{P+1}, \mathbf{d}'_{P+2}, \cdots, \mathbf{d}'_K$ are components with lower reproducibility, which may be truncated as the components due to noise.

The dimension $P$ of the principal basis in Eqn.10 could be simply decided according to the histogram of $\{\|\mathbf{\lambda}'_k\|_0\}_{k=1}^K$. One can set the maximum point of the histogram to $P$:

$$P = \arg\max_k Hist(\|\mathbf{\lambda}'_k\|_0) \tag{11}$$

Fig.2 illustrates this systematic procedure of determining threshold of reproducibility $P$. A sparse decomposition works on a noisy image (Fig2a), with the peak signal-to-noise ratio about $15dB$ in this example. Setting $K = 256$, we show the $256$ atoms (patches here) in Fig.2b and the corresponding $\ell^0$-norms $\{\|\mathbf{\lambda}_k\|_0\}_{k=1}^{256}$ in Fig.2d where the red line should be chosen as a suitable threshold $P$ of $\|\mathbf{\lambda}\|_0$, which means that those atoms with higher reproducibility than this threshold shall be used to constitute a principal signal basis. Alternatively, sorting $\{\|\mathbf{\lambda}_k\|_0\}_{k=1}^{256}$ in descending order into $\{\|\mathbf{\lambda}'_k\|_0\}_{k=1}^{256}$ (Eqn.8), the number $P$ is just the maximum point of the histogram for $\{\|\mathbf{\lambda}'_k\|\}_{k=1}^K$ shown in Fig.2e, so that the first $P$ patches in the reordered dictionary $\mathbf{D}'$ (Fig.2c) compose a principal basis $\overline{\mathbf{D}}_P$ (Eqn.10).

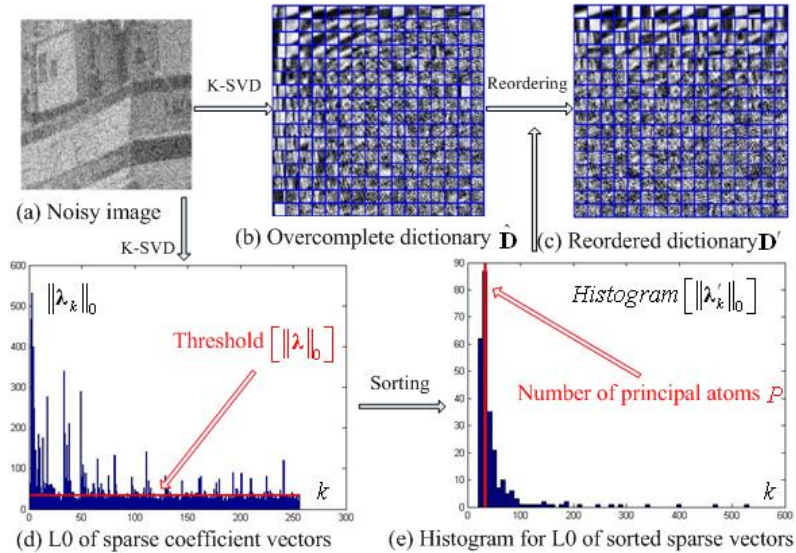

Figure 2. Systematic method for determining of number of principal atoms *P* according to sorted sparse vectors.

This proposed principal basis analysis (PBA) method (Eqns.5-6, Eqn.8 and Eqns.10-11) is particularly convenient to construct sparse signal approximations for applications such as noise removal, pattern recognition, machine learning, and so on.

Specifically, we want, from a corrupted data set $\mathbf{X} = \{\mathbf{x}_m\}_{m=1}^{M}$, to search underlying signal pattern $\mathbf{S}$. We shall construct an optimal approximation $\hat{\mathbf{S}}$ with the proposed principal dictionary $\bar{\mathbf{D}}_P$ by solving the following forms with the orthonormal matching pursuit (OMP) algorithm [6]:

$$\hat{\boldsymbol{\alpha}}_m = \arg\min_{\boldsymbol{\alpha}_m} \left\{ \mu_m \|\boldsymbol{\alpha}_m\|_0 + \|\bar{\mathbf{D}}_P \boldsymbol{\alpha}_m - \mathbf{x}_m\|_2^2 \right\}, \quad 1 \le m \le M$$

$$\hat{\mathbf{S}} = \arg\min_{\mathbf{S}} \left\{ \gamma \|\mathbf{S} - \mathbf{X}\|_2^2 + \sum_m \|\bar{\mathbf{D}}_P \hat{\boldsymbol{\alpha}}_m - \mathbf{x}_m\|_2^2 \right\}$$

(12)

where $\gamma$ is Lagrange multiplier and $\mu_m$s are parameters in the OMP algorithm.

## 4. APPLICATION OF PBA TO DENOISING

A major difficulty of denoising is to separate underlying signal details from noise. The proposed principal basis analysis (PBA) selects important components in sparse decomposition according to their reproducibility degree which will be rather high for signal details, edges or textures and so on; but very low for different kinds of white noises, Gaussian or non-Gaussian. The application of PBA to image denoising is here presented.

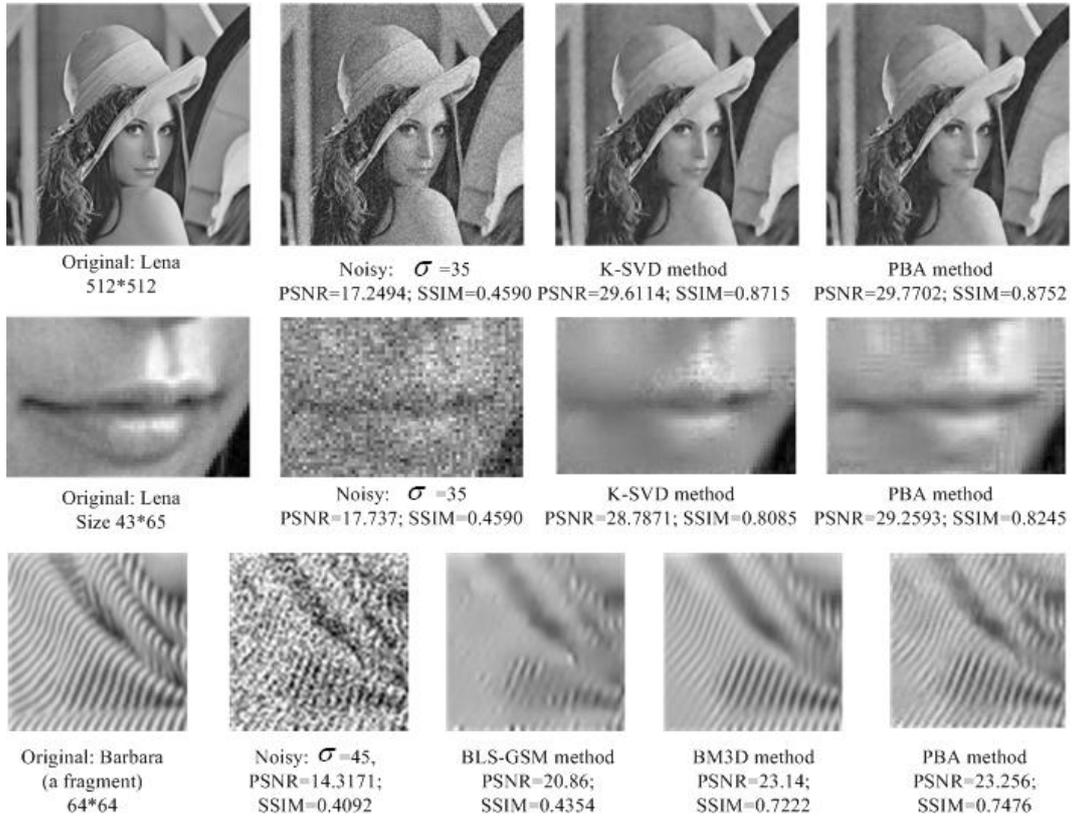

Figure 3. Comparison of image denoising between the proposed PBA method (the right column) and the K-SVD, BLS-GSM, BM3D methods in terms of PSNR and SSIM.

In this application, PBA method intends to preserve faint signal details under a situation of strong noise. We use the peak signal-to-noise ratio (PSNR) to assess the noise removal performance, and the structural similarity index metric (SSIM) between denoised image $\hat{\mathbf{S}}$ and the pure one $\mathbf{S}$ to evaluate the preserving details performance. In our experiments, dictionaries used $\mathbf{D}$s are of size $64 \times 256$ ($K = 256$ atoms), designed to handle image patches $\mathbf{x}_m$ of size pixels $N = 64 = 8 \times 8$.

In a first group of experiments, we consider noisy images $\mathbf{X} = \mathbf{S} + \mathbf{V}$ with an additive zero-mean white Gaussian noise $\mathbf{V}$. The standard deviation of noise is set to $\sigma = 35$ for Lena image and to $\sigma = 45$ for Barbara image. We compare PBA method with K-SVD method [7], BLS-GSM method [8] and BM3D method [9] which are some of the best denoising methods reported in the recent literatures.

From the results shown in Fig.3, PBA method outperforms K-SVD method by about $0.1$ to $0.5 dB$ in PSNR and by about 1 to $35\%$ in SSIM (depending on how much details in images and how faint the details) for both images. In terms of subjective visual quality, from fragments of Lena image (the 2nd row) and Barbara image (the 3rd row) in Fig.3, we can see that PBA method is better at preserving fine details than the other three methods.

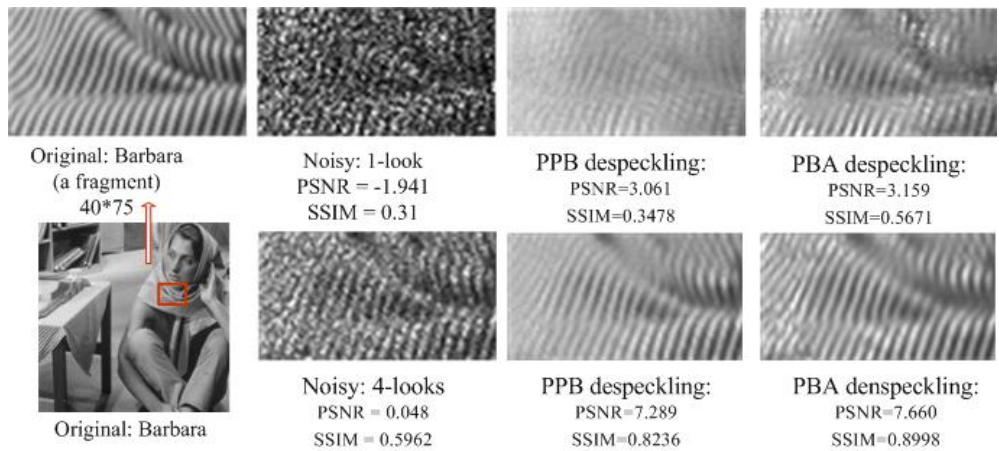

Figure 4. Comparison of image despeckling between the proposed PBA method and PPB method in terms of PSNR and SSIM.

In a second group of experiments, we consider simulated SAR images with speckle noise. Speckle is often modeled as a multiplicative noise as $\mathbf{X} = \mathbf{S} \bullet \mathbf{V}$ where $\mathbf{X}$, $\mathbf{S}$ and $\mathbf{V}$ correspond to the contaminated intensity, the original intensity, and the noise level, respectively. Taking multiplicative noise simulations, we considered one-look and 4-looks SAR scenario with a fragment of Barbara image.

We compare our PBA method with a probabilistic patch based (PPB) filter based on nonlocal means approach [10] which can cope with non-Gaussian noise. Fig.4 shows sample results. We can see, from the 3rd column in Fig.4, that PPB can well remove speckle noise; however, it also removes fine and low-intensity details. In comparison, the 4th column in Fig.4 shows advantages of our PBA method at preserving fine details and at suppressing strong noise.

## 5. CONCLUSION

A central idea of the proposed principal basis analysis (PBA) is to define reproducibility of component as the criterion to construct a principal signal basis. The reproducibility of components indicates actually an intrinsic characteristic of signal pattern in corrupted data set. PBA method takes advantages of interpretable components with prototype decomposition and of available importance score with sparse coefficient matrix. PBA could be interpreted as a PCA (Principal Components Analysis) in sparse representation of the signal so it can be expected that PBA, like PCA, has vast prospect in pattern recognition, machine learning and other application domains.

## ACKNOWLEDGMENTS

The idea of the principal basis analysis presented here arises through a lot of deep discussions with Professor Henri Maître and Professor Florence Tupin at Telecom-ParisTech in France.